# Predicting the Stereoselectivity of Chemical Transformations by Machine Learning

Justin Li
Del Norte High School
San Diego, CA, USA
jlisd04@gmail.com

Dakang Zhang
Department of Chemistry
Brandeis University
Waltham, MA, USA
dkzhang@brandeis.edu

Yifei Wang
Department of Computer Science
Brandeis University
Waltham, MA, USA
yifeiwang@brandeis.edu

Christopher Ye
Lexington High School
Lexington, MA
ye.chris135@gmail.com

Hao Xu
Department of Chemistry
Brandeis University
Waltham, Massachusetts, USA
haohxu@brandeis.edu

Pengyu Hong
Department of Computer Science
Brandeis University
Waltham, Massachusetts, USA
hongpeng@brandeis.edu

*Abstract* — **Stereoselective reactions (both chemical and enzymatic reactions) have been essential for origin of life, evolution, human biology and medicine. Since late 1960s, there have been numerous successes in the exciting new frontier of asymmetric catalysis. However, most industrial and academic asymmetric catalysis nowadays do follow the trial-and-error model, since the energetic difference for success or failure in asymmetric catalysis is incredibly small. Our current understanding about stereoselective reactions is mostly qualitative that stereoselectivity arises from differences in steric effects and electronic effects in multiple competing mechanistic pathways. Quantitatively understanding and modulating the stereoselectivity of for a given chemical reaction still remains extremely difficult. As a proof of principle, we herein present a novel machine learning technique, which combines a LASSO model and two Random Forest model via two Gaussian Mixture models, for quantitatively predicting stereoselectivity of chemical reactions. Compared to the recent ground-breaking approach [1], our approach is able to capture interactions between features and exploit complex data distributions, which are important for predicting stereoselectivity. Experimental results on a recently published dataset demonstrate that our approach significantly outperform [1]. The insight obtained from our results provide a solid foundation for further exploration of other synthetically valuable yet mechanistically intriguing stereoselective reactions.**

## I. Introduction

Stereochemistry plays an essential role in biology and numerous biochemical processes in living cells rely on stereospecific or stereoselective reactions. Notably, proteins and complex carbohydrates synthesized in human bodies are predominantly composed of enantiomerically pure amino acids and monosaccharides. It is also well known in medicines that stereochemistry is important to drug action. For some therapeutics, single-stereoisomer formulations can lead to improved therapeutic indices because they provide greater selectivity for their biological targets and/or better pharmacokinetics than a mixture of stereoisomers. While one stereoisomer can have positive effects on the body, another stereoisomer may be less effective (D-Isoproterenol vs L-Isoproterenol on the blood pressures or heart rate), ineffective (as in the case of the R enantiomer of ibuprofen), or even toxic (as in the case of thalidomide).

Stereoselectivity, most notably enantioselectivity, is one of the most important yet intriguing aspects of synthetic chemistry. Stereoselectivity can vary greatly in degree depending on reactants, catalysts, and reaction conditions. Quantitatively understanding and controlling the stereoselectivity of a chemical transformation – the relative proportions in which a non-stereospecific chemical transformation generates different stereoisomers under varying reaction conditions – is thus hugely important for organic synthesis. Yet we have only the most basic, qualitative understanding of the stereoselectivity of chemical transformations. We know that the stereoselectivity arises from differences in steric effects and electronic effects in the mechanistic pathways, but it has been frustrating to try to accurately, quantitatively rationalize and even predict stereoselectivity. In addition, the optimizations of asymmetric transformations have been mainly by trial-error. A huge volume of data about the stereoselectivity of chemical transformation has been published over the past 100+ years, and volumes more are now generated. Machine learning has emerged as an effective avenue for taking advantage of these data to build computational models for accurately and quantitatively predicting the stereoselectivity of chemical transformation.

In [1], Reid and Sigman applied machine learning to predict the stereoselectivity of chiral phosphoric acid (CPA)-catalyzed addition of protic nucleophiles to imines in which chiral 1,1'-bi-2-naphthol(BINOL)-derived phosphoric acid catalysts bearing aromatic groups at the 3 and 3' seem effective for high enantioselectivity for a range of nucleophilic addition reactions. The CPA catalysts have been widely used in a variety of highly enantioselective reactions [2-3]. Therefore, the authors collected 381 published CPA reactions with varied reaction components from 17 sources [4-20], and generated a set of molecular features (both geometrical and topological) to describe each imine, nucleophile, catalyst and solvent. Linear regression models were trained to predict enantioselectivity using those features as well as other reaction variables (e.g., concentration of reagents or catalysts, inclusion of molecular sieves, etc.). Although linear regression models are straightforward to interpret, they fall short of capturing interactions between features. In addition, it turns out that the chosen CPA reaction family has a complex data distribution, which is beyond the capacity of merely one linear regression model.

Herein, we describe exploring other more complex machine learning methods (e.g., LASSO [21], regression tree [22], random forest (RF) [23], and boosting tree [24]), yet with the same dataset, to achieve the previously difficult goals, including feature selection, capture of interactions between features, and compatibility with complex data distributions. The insight obtained from these methods provides a solid foundation for further exploration of other synthetically valuable yet mechanistically intriguing stereoselective reactions. Finally, we develop a composite machine learning model that delivers significantly better performance.

## II. METHODS

### A. Data

The training set contains 381 CPA reactions obtained from [1]. Each reaction includes a substrate, solvent, catalyst, nucleophile, and imine. Numerical features were derived from DFT calculations and molecular topologies to describe solvents (160 properties), catalysts (85 properties), nucleophiles (15 properties), and imines (22 properties). The activation energy ($\Delta\Delta G^\ddagger$) and reaction variables of each reaction were also collected. Additionally, 64 out-of-sample reactions (i.e., test data) were also collected by [1] from 3 sources [25-27]. The goal is to build a robust model that predicts the $\Delta\Delta G^\ddagger$ value of a reaction given the properties of catalyst, imine, nucleophile, and solvent in this reaction.

### B. Stereoselectivity Prediction Model Development

We first tested four widely used machine learning techniques (i.e., LASSO [21], regression tree [22], random forest (RF) [23], and boosting tree [24]) for predicting $\Delta\Delta G^\ddagger$ values using all imine, nucleophile, catalyst, and solvent features. LASSO [21], or Least Absolute Shrinkage and Selection Operator, is a type of linear regression that uses the shrinkage technique to shrink the model's coefficients towards a central point. It performs variable selection by using the L1-regularization, which penalizes based on the magnitude of coefficients in order to bring coefficients closer to zero. As a result of this penalty, many coefficients will be brought to zero, and only the variables that are strongly associated with the response variable will remain to make predictions. In our study, the L1-regularization effectively produces a simpler, more interpretable model by selecting a small subset of important features to be used in describing linear relationships between input features and output ($\Delta\Delta G^\ddagger$ in our context).

Regression tree [22] is a type of non-parametric model which uses simple decision rules to make predictions. The model partitions data sets into smaller groups at a series of decision nodes, at which a certain path is followed depending on whether or not a condition is met. Regression tree is able to capture more complex, non-linear relationships between features in predicting $\Delta\Delta G^\ddagger$ values.

Random forest [23] is an ensemble learning method that assembles a collection of regression trees trained on various sub-samples of features of the training dataset. The resulting ensemble regressor reduces the overfitting and stability issues faced by one single regression tree. The prediction made by a random forest model is the mean output of its regression trees.

Similar to random forest, boosting tree [24] is another ensemble learning method that constructs a collection of regression trees to make predictions. However, different from a random forest model that builds independent trees and averages their results to produce predictions, boosting tree uses the boosting technique [30] to train and combine a series of trees in order to produce results better than those of individual trees. Each tree added by the boosting technique is trained to minimize the error of the previous tree. The boosting training procedure also generates a weight for each tree. The final prediction made by a boosting tree model is the weighted average of predictions made by individual trees.

We used the training set and performed 2-fold cross-validation to compare these four techniques. The results (Table 1) indicate that RF performed the best with the mean test $r^2$ value of 0.926 and the test mean squared error (MSE) value of 0.223. Fig. 1 illustrates the results of a typical 2-fold cross validation run. In [1], their linear regression model produced a test $r^2$ value of 0.87, and identified 6 important features: the imine properties "C", "N", and "L2", the nucleophile property "H-X-CNu", the catalyst properties "SubS", and the solvent property "Balaban-type index from polarizability weighted distance matrix". In our RF model, the top five most importance features contribute almost 80% of the total importance (Table 2) and include the imine property "C" and the nucleophile property "H-X-CNu". Nevertheless, different from the discoveries in [1], our RF model found that catalyst and solvent properties had small impacts on the overall prediction. A possible explanation is that the training data lacks variations in catalysts and solvents, i.e., the reactions in the training set share a few catalysts and solvents. We also ran fourfold cross validation with the training data using our RF model and received an average $r^2$ value of 0.975. In [1], their linear regression model had an average fourfold cross validation $r^2$ value of 0.87.

TABLE 1: Comparing LASSO, RT (regression tree), RF (random forest) and BT (boosting tree) for predicting stereoselectivity. The whole training set was used in two-fold cross-validation runs (100 times). The mean results and the corresponding standard deviation (STD) values are listed.

| Models | Test MSE (STD) | Test $r^2$ (STD) | Train $r^2$ (STD) |
|--------|---------------|------------------|-------------------|
| LASSO  | 0.343 (0.048) | 0.887 (0.018)    | 0.942 (0.008)     |
| DT     | 0.361 (0.083) | 0.880 (0.0278)   | 0.999 (0.001)     |
| BT     | 0.225 (0.040) | 0.925 (0.014)    | 0.987 (0.002)     |
| RF     | 0.223 (0.048) | 0.926 (0.014)    | 0.987 (0.002)     |

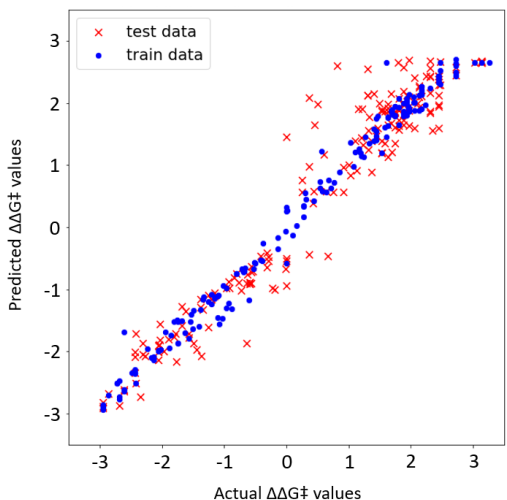

Figure 1: The results of RF in a typical 2-fold cross validation run using the training data. The blue dots are training samples, and the red dots represent the validation samples. The training and predicted $r^2$ values are 0.987 and 0.926 respectively

TABLE 2: The top five most important properties used by the RF model. All features are used. The **Feature** column lists the feature names. The **Molecule** column lists the molecule category of each feature. The **Importance** column list the importance weight of each feature.

| Feature | Molecule | Importance (out of 100) |
|---------|----------|-------------------------|
| C | imine | 54.59 |
| SL | imine | 17.26 |
| H-X-Nu | nucleophile | 2.56 |
| PG | imine | 2.16 |
| H-X-CNu | nucleophile | 1.85 |

Since the imine gives strong indications towards what the products of a CPA reaction are and it requires extra efforts to obtain imine information, we investigate the possibility of predicting the ΔΔG‡ values without the knowledge of imine features. Interestingly, better machine learning models (see Table 3) could be trained without using information about the imine. Again, RF performed the best with a mean test $r^2$ of 0.933 and a mean test MSE of 0.203. Fig. 2 illustrate the results of a typical 2-fold cross validation run. All top 5 most important features used by the RF model are nucleophile features (Table 4). Once again, catalyst features are only slightly influential on the final prediction. The most important catalyst properties include "iPOsy", "B1", and "C1", each of which has an importance of only around 0.5. In the rest of this paper, we use nucleophile-focused models to refer to these models trained without using imine features.

TABLE 3: Comparing LASSO, RT (regression tree), RF (random forest) and BT (boosting tree) for predicting stereoselectivity without using imine information. The training set was used. Two-fold cross-validation was run 100 times. The mean results and the corresponding standard deviation (STD) values are reported here.

| Models | Test MSE (STD) | Test $r^2$ (STD) | Train $r^2$ (STD) |
|--------|----------------|------------------|-------------------|
| LASSO | 0.625 (0.086) | 0.793 (0.033) | 0.864 (0.017) |
| DT | 0.291 (0.057) | 0.904 (0.020) | 0.977 (0.003) |
| BT | 0.234 (0.032) | 0.923 (0.011) | 0.967 (0.004) |
| RF | 0.203 (0.0345) | 0.933 (0.012) | 0.972 (0.004) |

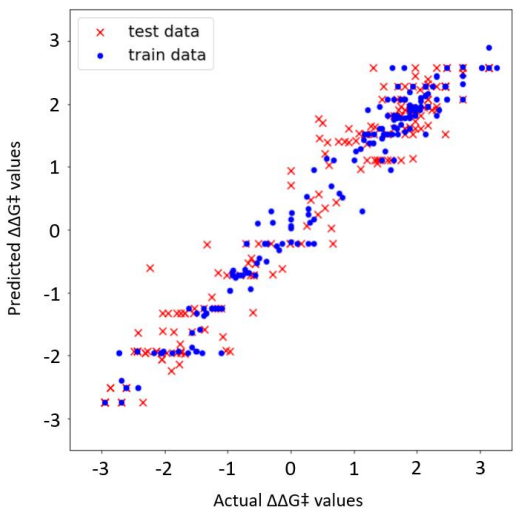

Figure 2: The results of the nucleophile-focused RF in a typical 2-fold cross validation run using the training data. The blue dots are training samples, and the red dots represent the validation samples. The training and predicted $r^2$ values are 0.972 and 0.933 respectively

TABLE 4: The top 5 most important properties used by random forest when the imine features are excluded. The **Feature** column lists the feature names. The **Molecule** column lists the molecule source of each feature. The **Importance** column list the importance weight of each feature.

| Feature | Molecule | Importance (out of 100) |
|---------|----------|-------------------------|
| H-X-Nu | nucleophile | 20.24 |
| H-X-CNu | nucleophile | 17.87 |
| Nu | nucleophile | 16.31 |
| Polarizability | nucleophile | 13.01 |
| iXH | nucleophile | 3.59 |

To compare with the nucleophile-focused models, we examined the performance of imine-focused models trained without using nucleophile properties (Table 5). Interestingly enough, the imine-focused models did not perform quite as well as our other models reported above. Random forest performed the best with a mean test $r^2$ of 0.881 and a mean test MSE of 0.360. Combining with the results in Tables 1 and 2, we hypothesize that imine properties can be well explained by the other molecules involved in the same reactions. Hence, we applied random forest to predict the imine transition state (i.e., E- or Z-imine) by using the features of catalyst, nucleophile, and solvent. The two-fold cross-validation results (training and test accuracies are 0.993 and 0.970, respectively) indicate that there is a strong link between the transition states and the corresponding nucleophilic reactants, which resonates the results of the nucleophile-focused RF model.

TABLE 5: Results of the imine-focused models excluding the nucleophile features

| Models | Test MSE (STD) | Test $r^2$ (STD) | Train $r^2$ (STD) |
|---|---|---|---|
| Lasso | 0.626 (0.120) | 0.794 (0.040) | 0.874 (0.014) |
| DT | 0.524 (0.100) | 0.827 (0.036) | 0.975 (0.005) |
| BT | 0.376 (0.041) | 0.876 (0.015) | 0.966 (0.006) |
| RF | 0.360 (0.045) | 0.881 (0.015) | 0.966 (0.007) |

We observed that the training results are noticeably better than the test results in the two-fold cross validation experiments. This could be problematic when applying the above models to new CPA reactions whose imines and/or nucleophiles are very different from those in the training data (i.e., new samples could fall in the low-density regions of the training data). In such a scenario, a simpler model (e.g., linear regressor), which assumes less about data distribution and requires less amount of training data, may deliver better extrapolation than RF. One intriguing solution is to train multiple prediction models and combine them into a composite model. Based on our observations described above, we decided to include three prediction models in this composite model: a random forest model trained by all features (overall RF model), a second random forest model trained without imine features (nucleophile-focused RF model), and a linear regression model trained via the LASSO algorithm using all features. We chose not to include the imine-focused RF model (i.e., random forest model trained without nucleophile features) due to its relatively poorer performance compared to other models. The overall RF model is able to make strong predictions when both the imine and nucleophile of a reaction are similar to those in the training data. The nucleophile-focused RF model is able to make strong predictions when the nucleophile of a reaction is similar to those in our training data while the imine is not. The LASSO model assumes the least about the data and may extrapolate relatively more effectively when the nucleophile of a reaction is not similar to those in our training data.

Eventually, we decided to build a composite model with an architecture shown in Fig. 3. Given a new sample, the composite model first estimates its probability density values with respect to the nucleophile and imine distributions in the training data, and then chooses a prediction model accordingly.

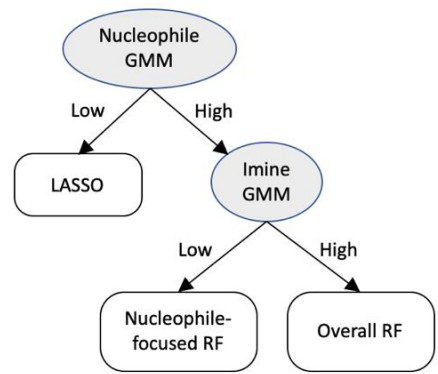

Figure 3. The composite model estimate the probability density values of a reaction with respect to the distributions of nucleophiles and imines, which are approximated by the nucleophile GMM and the imine GMM, respectively. One of three models (LASSO, nucleophile-focused RF, and overall RF) is then chosen accordingly to predict the activation energy (ΔΔG‡) of the given reaction.

The nucleophile and imine density functions are respectively approximated by two Gaussian mixture models (GMM) [28] fitted on the training data via the EM algorithm [29]. The nucleophile GMM approximates the joint distribution of important nucleophile features (H-X-Nu, H-X-CNu, Nu, and Polarizability) chosen as top important features by the nucleophile-focused RF model (Table 4). The imine GMM approximate the joint distribution of the iminium features (C, SL, and PG) because of their importance in the overall random RF model (see Table 2). The nucleophile and imine GMMs contain 14 and 15 gaussian components, respectively, which were decided based on their BIC scores in training. A density value is considered high if it is larger than 1, otherwise low. The above setting was validated in a leave-one-reaction-type-out cross-validation (i.e., use 16 reaction types to train the composite model and test it on the left-out reaction type), the composition model outperformed individual models (i.e., overall RF, nucleophile-focused RF, and LASSO). Its mean absolute error per reaction type is on average 0.05 less than the overall RF, 0.23 less than the nucleophile-focused RF, and 0.11 less than LASSO.

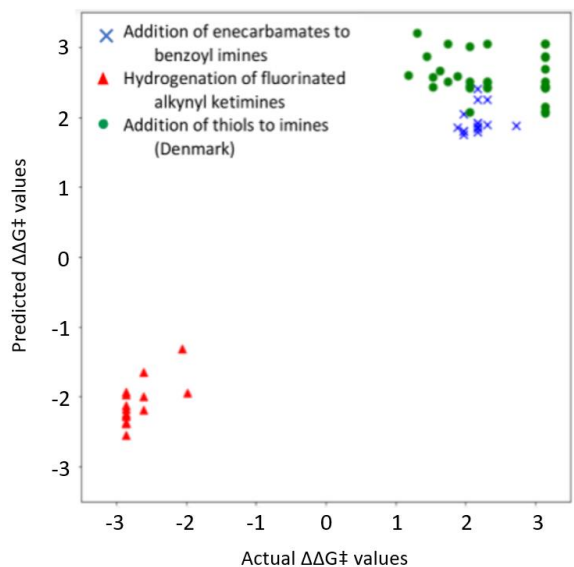

Figure 4. The test performance of the composite model

TABLE 6: Test results of the composite model. The 1st row lists three reaction types used in test. The 2nd row shows the predictors chosen by the composite model for the corresponding reaction types listed in the 2nd, 3rd, and 4th columns. The third row lists the mean log(probability density value) of each reaction type with respect to imine-GMM. The fourth row lists the mean log(probability density value) of each reaction type with respect to nucleophile-GMM. The third and fourth rows detail how the composite model chooses a predictor for each reaction type. The fifth row lists the results of the composite model. The last row lists results reported in [1]. The numbers outside of the parentheses are the results of the comprehensive model trained by using all training data. The results inside the parentheses were obtained by the E-/Z-imine models trained by using the training reactions that proceed via the E-/Z-imine transition states, respectively.

| Reaction types | Addition of enecarbamates to benzoyl imines [25] | Hydrogenation of fluorinated alkynyl ketimines [26] | Addition of thiols to imines (Denmark) [27] |
|---|---|---|---|
| Predictor chosen by the composite model | LASSO | Nucleophile-focused RF | Overall RF |
| Average log(imine-GMM density value) | 8.43 | -931.35 | 8.90 |
| Average log(nucleophile-GMM density value) | -646434.250 | 14.82 | 10.60 |
| Mean Absolute Error (MAE) | 0.25 | 0.26 | 0.52 |
| MAE in [1] | 0.37 (0.24) | 1.0 (0.30) | 0.65 (0.67) |

We also tested the composite model on out-of-sample data not included in the training data. In [1], 64 out-of-sample reactions were also collected: 15 reactions from "addition of enecarbamates to benzoyl imines" [25], 15 reactions from "hydrogenation of fluorinated alkynyl ketimines" [26], and 34 reactions from "addition of thiols to imines (Denmark)" [27]. These test samples were also used in our experiment, and the results are summarized in Table 6 and visualized in Fig. 4. The composite model chose LASSO for the reactions in the "addition of enecarbamates to benzoyl imines" type and produced a mean absolute error (MAE) of 0.25, compared to 0.37 by the comprehensive model and 0.24 by the E-imine model in [1]. The nucleophile-focused RF model was chosen for the 15 reactions in the "hydrogenation of fluorinated alkynyl ketimines" category, which led to a MAE of 0.26, compared to 1.0 by the comprehensive model and 0.30 by the Z-imine model in [1]. Finally, the overall RF model was chosen for the 34 reactions in the "addition of thiols to imines (Denmark)" category to produce a MAE of 0.52, compared to 0.65 by the comprehensive model and 0.67 by the E-imine model in [1]. Across all 64 out-of-sample predictions, the MAE of our composite model is 0.39 and the $r^2$ value is 0.951, demonstrating the generalizability of our approach.

## III. CONCLUSION AND DISCUSSIONS

Using the CPA reaction family as the proof-of-the-principle case, we developed a novel composite machine learning model that could learn from existing stereoselective reactions and then accurately and quantitatively predicted the activation energy of new reaction. The inputs of the composite model are features extracted from molecules involved in reactions. The composite model is able to explore interaction between molecular features in addition to selecting features most relevant to making predictions. The features selected by our model could give us better understandings about stereoselectivity reactions and may enable better ways to optimize certain reactions. The model appreciates the complex landscape of stereoselective reactions and uses GMMs to approximate the distributions of key molecules (nucleophile and imine), which allows the composite model to select more appropriate predictors and take advantage of the strengths of different machine learning models.

The imine molecules represent the transition states of stereoselective reactions and are essential to human understanding of the stereoselectivity. To our surprise, within the context of the training data used in this study, we found that the properties of imine are not indispensable in accurately predicting the activation energy of stereoselective reactions. This may be due to the specific distribution of our training data. Another explanation is imine features can be well explained by other molecular features, especially, nucleophile features, which is coupled with our finding that the transition state (i.e., E-/Z-imine) can be almost perfectly predicted by using other molecules. We also found that catalyst features are less important for making predictions. This may be because the structures of the catalysts used by the CPA reactions are similar. Finally, the generalizability demonstrated by our approach in test suggests that it can be applied to other stereoselective reaction families. We expect to discover novel insights into catalysts when integrating data of diverse stereoselective reaction families.